\documentclass[journal]{IEEEtran}

\usepackage{amsmath}            
\usepackage[figuresright]{rotating}
\usepackage{subfigure}
\usepackage{booktabs} 
\usepackage{graphicx}
\usepackage{color,soul}
\usepackage{amsmath, amsthm, amssymb}
\usepackage{framed}
\usepackage{float}
\usepackage{url}
\usepackage{fancyhdr}
\usepackage{algorithm}
\usepackage{pifont}
\usepackage[noend]{algpseudocode}
\usepackage{listings}
\usepackage{tikzpagenodes}

\begin{document}

\title{Cyber-All-Intel: An AI for Security related Threat Intelligence}

\author{\IEEEauthorblockN{
Sudip Mittal,
Anupam Joshi and 
Tim Finin
}\\
\IEEEauthorblockA{University of Maryland, Baltimore County, Baltimore, MD 21250, USA\\ Email: $\lbrace$smittal1,joshi,finin$\rbrace$@umbc.edu}
}

\maketitle
\begin{abstract}
Keeping up with threat intelligence is a must for a security analyst today. There is a volume of information present in `the wild' that affects an organization. We need to develop an artificial intelligence system that scours the intelligence sources, to keep the analyst updated about various threats that pose a risk to her organization. A security analyst who is better `tapped in' can be more effective.

In this paper we present, {\em Cyber-All-Intel} an artificial intelligence system to aid a security analyst. It is a system for knowledge extraction, representation and analytics in an end-to-end pipeline grounded in the cybersecurity informatics domain. It uses multiple knowledge representations like, vector spaces and knowledge graphs in a `VKG structure' to store incoming intelligence. The system also uses neural network models to pro-actively improve its knowledge. We have also created a query engine and an alert system that can be used by an analyst to find actionable cybersecurity insights. 

\end{abstract}

\section{Introduction}
In the broad domain of security, analysts and policy makers need knowledge about the state of the world to make critical operational/tactical as well as strategic decisions. One such source of knowledge is threat intelligence. It plays a vital role in helping a security analyst mount a defense and is nearly always dependent on global events; for example, consider the following timeline: 

\begin{itemize}
\item Microsoft publishes exploit details and issues patches on March 14, 2017 for a critical vulnerability which allows remote code execution if an attacker sends specially crafted messages to a Microsoft Server Message Block 1.0 (SMBv1) server \cite{ms017bulletin}.
\item An exploit named ETERNALBLUE is leaked by the `Shadow Brokers' hacker group on April 14, 2017 affecting various versions of the Microsoft Windows operating system \cite{BBCshadow}. It uses the same vulnerability mentioned above. 
\item ETERNALBLUE was used during the {\em WannaCry} ransomware attack on May 12, 2017 \cite{wiredwannacry}.
\item It was also exploited to carry out the {\em NotPetya} cyberattack on June 27, 2017 \cite{nytnotpetya}.
\end{itemize}


When we consider this example from the security analyst point of view, she may not be aware of recent `intelligence' available in `the wild' and/or may be lazy/slow in updating her security configurations.
What we need is an artificial intelligence based system that aids the security analyst. Such an AI should strive to keep the analyst updated and also issue timely alerts. The work in this paper, aims to prototype a system that can scour OSINT sources for such information and make them accessible to an analyst. The better ``tapped in'' the analyst is to the potential threat landscape, the better they are able to detect attacks.

In modern enterprises, security analysts monitor threats in a security operations center (SoC) by {\em watchstanding}, akin to a lookout on a ship watching the environs for danger. Screens typically show warnings and alerts from individual products and detectors that the enterprise has installed. Watchstanding permits a highly trained security analyst to look at all the disparate pieces of information, and see if they `click together' to form some pattern which might indicate an attack. 

The detection efficacy of a security analyst depends on her operational and strategic knowledge about current security landscape and the associated intelligence. This enables her to better interpret the data from the Security Information and Event Management (SIEM) systems in the SoC. Specifically, the analyst is aided by her background knowledge regarding the context of the system (e.g., what kinds of applications are installed on their system, what the systems normal behavior pattern is, what potential vulnerabilities might exist, what information might an adversary be after?), and the external world (e.g., ``intelligence'' about what new attacks that exist in the wild or are being discussed as possibilities, hacktivists discussing attacking a particular country or organization, etc.). Unfortunately, the knowledge about these security vulnerabilities and planned attacks is scattered on the dark web vulnerability markets, user/product forums, social media services, blogs, etc. However, even the best of SIEM systems today do not effectively reason on ``intelligence" about the state of the cyberworld, such as an analyst might obtain by talking to peers or by looking, for instance, at dark web updates, security blogs, etc. 


In this paper we present, a cyber security informatics system - `{\em Cyber-All-Intel}'. The system takes as input cybersecurity related text data from various unstructured sources like Dark Web, blogs, social media, National Vulnerability Databases (NVDs), newspaper articles, etc. and represent the extracted knowledge in the `{\em VKG structure}'. We extract, represent and integrate the knowledge present in a variety of Open Source Intelligence (OSINT) web fora as entities, then use the resulting knowledge graph and embeddings to obtain actionable cybersecurity information for the analyst.

The system is built using the VKG structure -- a hybrid structure that combines knowledge graph and embeddings in a vector space. The structure creates a new representation for relations and entities of interest. In the VKG structure, the knowledge graph includes explicit information about various entities and their relations to each other grounded in an ontology\footnote{\url{https://www.w3.org/standards/semanticweb/ontology}}. The vector embeddings, on the other hand, include implicit information found in context where these entities occur in a corpus. The base ontology is enhanced to have relations that describe the vector embeddings associated with terms in the ontology.

The Cyber-All-Intel system also pro-actively tries to improve the underlying cybersecurity knowledge. The vector part of the VKG structure is used to improve the knowledge graph part and vice versa. We utilize powerful deep neural networks to automatically update the underlying knowledge. Such an ability allows the system to be more accurate and best assist the security analyst in her tasks. 

We have also included two applications in the Cyber-All-Intel system, namely, an alert recommender and a query processing engine that leverage the advantages provided by the VKG structure. The security analyst can ask the Cyber-All-Intel system to issue alerts based on an organization's `{\em system profile}'. She can dig deeper into an alert by asking complex queries like, `{\em Raise an alert if, a vulnerability similar to denial of service is listed in MySQL}' and get an answer from the system. 

Major contributions presented in this paper include: 
\begin{itemize}
\item Populating unstructured cybersecurity knowledge in Vectorized Knowledge Graphs (VKG) and creating the Cyber-All-Intel system.
\item Cybersecurity Knowledge Representation Improvement: We aim to use the vector space to improve the knowledge graph representation and the knowledge graph to improve the vector space representation (See Section \ref{imp}).
\item Actionable Cybersecurity Insights from Vectorized Knowledge Graphs:  we will create agents that fully utilize the advantages provided by VKGs (Section \ref{adv}) for query processing (Section \ref{query}), generating alerts for threats (Section \ref{rec}), and finding similar attacks. 
\end{itemize}

The rest of the paper is as follows -- In Section \ref{relwork}, we discuss the related work and some background on knowledge representation. We describe the Cyber-All-Intel system architecture and pipeline in Section \ref{system}. Section \ref{model} gives various details about the VKG structure. We discuss cybersecurity knowledge improvement in Section \ref{imp}. The query processing and alert generation applications have been discussed in Section \ref{appl}. We present our evaluation in Section \ref{eval}, we conclude in Section \ref{conc}.

\section{Background \& Related Work}\label{relwork}

In this section we present some background and related work in the field of knowledge graphs, vector space models, text extraction, and knowledge representation in cybersecurity. 

\subsection{Knowledge Representation for Cybersecurity}

Knowledge graphs have been used in cybersecurity to combine data and information from multiple sources. Undercofer et al. created an ontology by combining various taxonomies for intrusion detection \cite{Undercoffer2003b}. Kandefer et al. \cite{kandefer2007symbolic} created a data repository of system vulnerabilities and with the help of a systems analyst, trained a system to identify and prevent system intrusion. Takahashi et al. \cite{takahashi2010ontological,takahashi2010building} built an ontology for cybersecurity information based on actual cybersecurity operations focused on cloud computing-based services.  Rutkowski et al. \cite{rutkowski2010cybex} created a cybersecurity information
exchange framework, known as CYBEX. The framework describes how
cybersecurity information is exchanged between cybersecurity entities
on a global scale and how implementation of this framework will
eventually minimize the disparate availability of cybersecurity
information. Another insightful work by Xie et al. \cite{xie2010using} discusses uncertainty modeling for cyber security centered around near real-time security analysis such as intrusion response. In this paper the authors use Bayesian networks to model uncertainty in enhanced security analysis.

In our previous work, Syed et al. \cite{syed2015uco} created the Unified Cybersecurity Ontology (UCO) that supports information integration and cyber situational awareness in cybersecurity systems. The ontology incorporates and integrates heterogeneous data and knowledge schema from different cybersecurity systems and most commonly-used cybersecurity standards for information sharing and exchange such as STIX \cite{barnum2012standardizing} and CYBEX \cite{rutkowski2010cybex}. The UCO ontology has also been mapped to a number of existing cybersecurity ontologies as well as concepts in the Linked Open Data cloud.

\subsection{Text Extraction for Cybersecurity}

In our various preliminary systems \cite{Joshi-ICSC-2013,mittal2016cybertwitter} we demonstrate the feasibility of automatically generating RDF linked data from vulnerability descriptions collected from the National Vulnerability Database \cite{NVD}, Twitter \cite{twitter}, etc. Joshi et al. \cite{Joshi-ICSC-2013} extract information on cybersecurity-related entities, concepts and relations which is then represented using custom ontologies for the cybersecurity domain and mapped to objects in the DBpedia knowledge base \cite{auer2007dbpedia} using DBpedia Spotlight \cite{mendes2011dbpedia}. CyberTwitter \cite{mittal2016cybertwitter}, a framework to automatically issue cybersecurity vulnerability alerts to users. CyberTwitter converts vulnerability intelligence from tweets to RDF. It uses the UCO ontology (Unified Cybersecurity Ontology) \cite{syed2015uco} to provide their system with cybersecurity domain information.

\subsection{Vector Space Models \& Knowledge Graphs}

Extracting data from unstructured text (web) data sources, representing it, and reasoning over the representation to extract knowledge and information is one of the central challenges in the field of Artificial Intelligence. In addition to information extraction, it involves designing representations that capture the extracted information and that can be used to analyze it. There is an inherent information loss while representing knowledge through different methods. Consider two representations that are heavily used in literature -- Knowledge Graphs and Vector Space Embeddings. By representing knowledge as vector embeddings, we lose the explicit declarative character of the information. Knowledge graphs on the other hand are adept at asserting declarative information, but miss important contextual information around the entity or are restricted by the expressibility of the baseline ontology used to represent the knowledge \cite{davis1993knowledge}.

It is important to highlight that both of the knowledge representation techniques provide applications built on these technologies certain advantages. Embeddings provide an easy way to search their neighborhood for similar concepts and can be used to create powerful deep learning systems for specific complex tasks. Knowledge graphs provide access to versatile reasoning techniques. Knowledge graphs also excel at creating rule-based systems where domain expertise can be leveraged. To overcome limitations of both and take advantage of their complementary strengths, {\em we propose the VKG structure that is part knowledge graph and part vector embeddings} (Section \ref{model}). VKG is more than the sum of these parts and can be used to develop powerful inference methods and a better semantic search. 

Word embeddings are used to represent words in a continuous vector space. Two popular methods to generate these embeddings based on `Relational Concurrence Assumption' are word2vec \cite{mikolov2013distributed,mikolov2013efficient} and GloVe \cite{pennington2014glove}. The main idea behind generating embeddings for words is to say that vectors close together are semantically related. Word embeddings have been used in various applications like machine  translation \cite{sutskever2014sequence}, improving local and global context \cite{huang2012improving}, etc. 

Modern knowledge graphs assert facts in the form of ($Subject$, $Predicate$, $Object$) triples, where $Subject$ and $Object$ are modeled as graph nodes and the edge between them ($Predicate$) model the relation between the two. DBpedia \cite{auer2007dbpedia}, YAGO (Yet Another Great Ontology) \cite{suchanek2007yago}, YAGO2 \cite{hoffart2013yago2}, Google Knowledge Graph \cite{officialgoogle2012}, etc. are some of the examples of popular knowledge graphs. 

An important task on both vector space models and knowledge graphs is searching for similar entities, given an input entity. In vector spaces, embeddings close together are semantically related and various neighborhood search algorithms \cite{gionis1999similarity,Kuzi:2016:QEU:2983323.2983876} have been suggested. On the other hand semantic similarity on knowledge graphs using ontology matching, ontology alignment, schema matching, instance matching, similarity search, etc. remains a challenge \cite{shvaiko2013ontology,DeVirgilio:2013:SMA:2457317.2457352,zheng2016semantic}. In this paper we use the VKG structure, in which we link the knowledge graph nodes to their embeddings in a vector space (see Section \ref{model}).  

Yang et al. \cite{yang2016fast} argued that a fast top-k search in knowledge graphs is challenging as both graph traversal and similarity search are expensive. The problem will get compounded as knowledge graphs increase in size. Their work proposes STAR, a top-k knowledge graph search framework to find top matches to a given input. Damljanovic et al. \cite{damljanovic2011random} have suggested using Random Indexing (RI) to generate a semantic index to an RDF graph \cite{rdf}. These factors combined have led to an increased interest in semantic search, so as to access RDF data using Information Retrieval methods. We argue that vector embeddings can be used to search, as well as index entities in a knowledge graph. We have built a query engine on top of the VKG structure that removes the need to search on the knowledge graph and uses entity vector embeddings instead (see Section \ref{model} and \ref{eval}). However, queries that involve listing declarative knowledge and reasoning are done on the knowledge graph part of the VKG structure. 


Vectorized knowledge graphs have also been created, systems like HOLE (holographic embeddings) \cite{DBLP:journals/corr/NickelRP15} and TransE \cite{wang2014knowledge} learn compositional vector space representations of entire knowledge graphs by translating them to different hyperplanes. Our work is different from these models as we keep the knowledge graph part of the VKG structure as a traditional knowledge graph so as to fully utilize mature reasoning capabilities and incorporate the dynamic nature of the underlining corpus for our cybersecurity use-case. Vectorizing the entire knowledge graph part for a system like Cyber-All-Intel will have significant computational overhead because of the ever-changing nature of vulnerability relations and velocity of new input threat intelligence.

In another work thread different from ours, vector models have also been used for knowledge graph completion. Various authors have come up with models and intelligent systems to predict if certain nodes in the knowledge graphs should have a relation between them. The research task here is to complete a knowledge graph by finding out missing facts and using them to answer path queries  \cite{lin2015learning,neelakantan2015compositional,socher2013reasoning,guu2015traversing}.

\begin{figure}[!htbp]
\centering
\includegraphics[width=\linewidth]{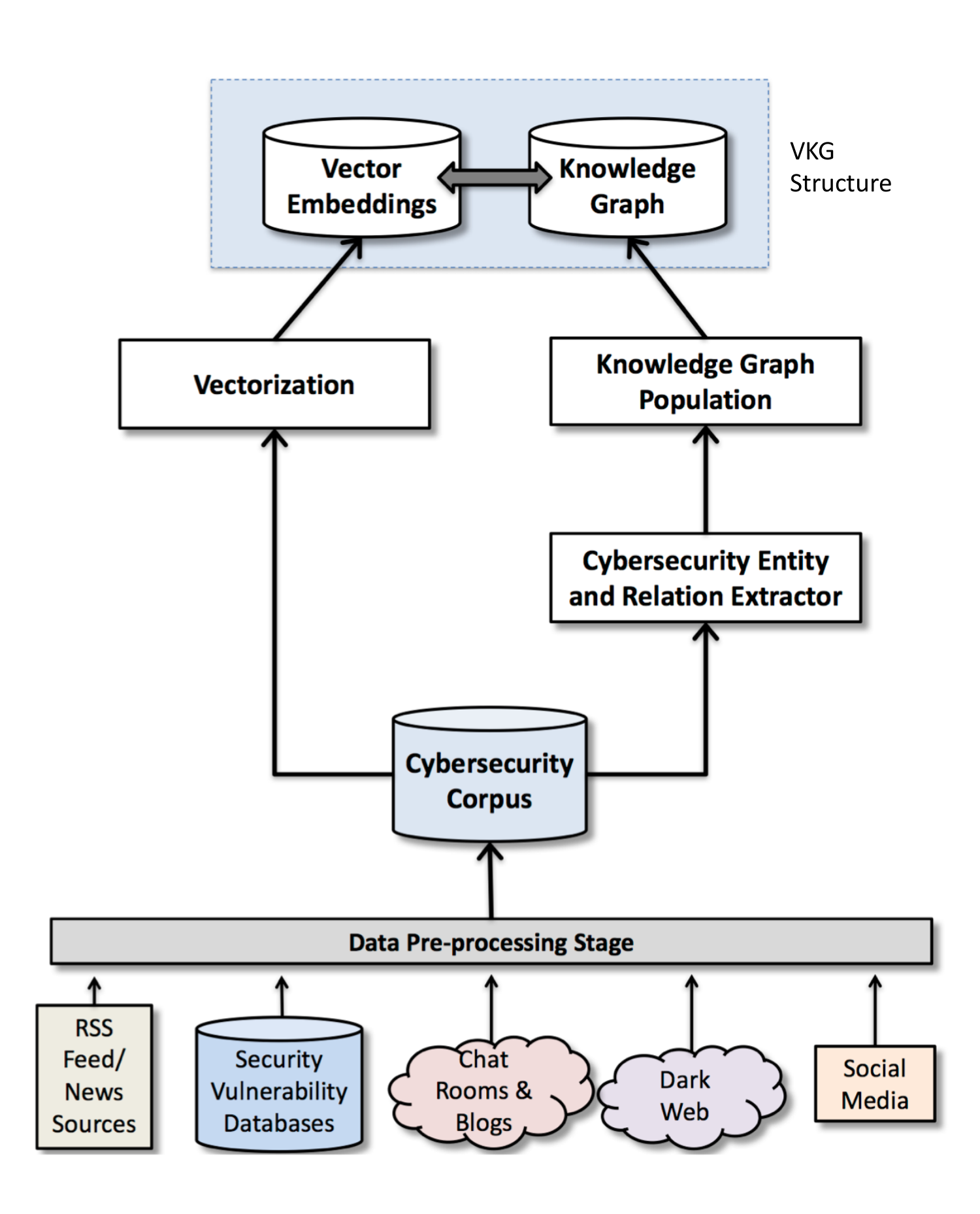}
\caption{C\MakeLowercase{yber}-A\MakeLowercase{ll}-I\MakeLowercase{ntel} System Architecture.}
\label{fig:arch}
\end{figure}

\section{C\MakeLowercase{yber}-A\MakeLowercase{ll}-I\MakeLowercase{ntel} System}\label{system}

In this section we discuss the overarching design and system architecture for Cyber-All-Intel (Figure \ref{fig:arch}). We first discuss various intelligence sources that serve as input to Cyber-All-Intel; then we go through  the architecture and the rationale behind our design decisions. Later we explain the knowledge representation techniques (VKG structure) used in our system along with its advantages. We also discuss a few agents that can leverage these representation techniques to provide value to a security analyst. 


\subsection{Cybersecurity Sources}

OSINT is intelligence gathered from publicly-available \emph{overt} sources such as newspapers, magazines, social-networking sites, video sharing sites, wikis, blogs, etc. In cybersecurity domain, information available through OSINT can compliment data obtained through traditional security systems and monitoring tools like Intrusion Detection and Prevention Systems (IDPS) \cite{more2012knowledge}. Cybersecurity information sources can be divided into two abstract groups, formal sources such as NIST's National Vulnerability Database (NVD), United States Computer Emergency Readiness Team (US-CERT), etc. and various informal sources such as blogs, developer forums, chat rooms and social media platforms like Twitter, Reddit \cite{reddit} and Stackoverflow \cite{stackoverflow}, these provide information related to security vulnerabilities, threats and attacks. A lot of information is published on these sources on a daily basis making it nearly impossible for a human analyst to manually comb through, extract relevant information, and then understand various contextual scenarios in which an attack might take place. A manual approach even with a large number of human analysts would neither be efficient nor scalable. Automatically extracting relevant information from OSINT sources thus has received attention from the research community~\cite{neri2009mining,OSINT2009,cluo2013}. 

An organization may also have access to various forms of propriety or \emph{covert} data sources. These data sources can also be added as modules, to our Cyber-All-Intel system. However, in this paper we only describe open data sources.

\subsection{System Pipeline \& Architecture}

Our system pipeline includes a data collection engine that pushes new data into the system from a multitude of data sources. 
The Cyber-All-Intel system (Figure \ref{fig:arch}) automatically accesses data from some of these sources like NIST's National Vulnerability Database, Twitter, Reddit, Security blogs, dark web markets \cite{dnmArchives}, etc. The system begins by collecting data in a modular fashion from these sources. Followed by a data pre-processing stage where we remove stop words, perform stemming, noun chunking, etc. The data is then stored in a cybersecurity corpus. 

After creating a cybersecurity corpus we use a Security Vulnerability Concept Extractor (SVCE) which extracts security related entities and understands their relationships. Our current SVCE \cite{Lal-MS-2013,Joshi-MS-2013}, trained using various natural language processing techniques, enables us to extract cybersecurity related terms from text, which can then be stored in our knowledge graph. The data is then tagged and vectorized. The tagged entities are then converted to their vector embeddings. These embeddings are then included in the knowledge graph. For more details on the VKG structure see Section \ref{model}.




The data is asserted in RDF using the Unified Cybersecurity Ontology (UCO) \cite{syed2015uco}. Ontologies like UCO, Intelligence \cite{mittal2016cybertwitter}, DBpedia \cite{auer2007dbpedia}, YAGO \cite{suchanek2007yago} have been used to provide cybersecurity domain knowledge. The vector part on the other hand was created using a vector generation algorithm (Section \ref{kgtovec}). An example is shown in Figure \ref{fig:cyberexample}, where we create the VKG structure for the textual input: 

\vspace{1mm}
\textit{Microsoft Internet Explorer allows remote attackers to execute arbitrary code or cause a denial of service (memory corruption) via a crafted web site, aka ``Internet Explorer Memory Corruption Vulnerability.''}
\vspace{1mm}

Triples generated for the above mentioned input text are shown in the Figure \ref{fig:exampleRDF}. 

The knowledge part of the VKG structure is completed once the triples are added to the system. We used Apache Jena \cite{jena} to store our knowledge graph. For example, in Figure \ref{fig:cyberexample}, once added the nodes are linked to the vector embeddings for `Microsoft\_Internet\_Explorer', `remote\_attackers', `execute\_arbitrary\_code', `denial\_of\_service', and `crafted\_web\_site'. For our system, we retrain the vector model every two weeks to incorporate the changes in the corpus. We give various details about system execution and evaluation in Section \ref{setup} and Section \ref{eval} respectively.

\begin{figure}[!htbp]
\small
\begin{minipage}{1\columnwidth} 

@prefix uco: $<$http://accl.umbc.edu/ns/ontology/uco\#$>$ . \\
@prefix intel: $<$http://accl.umbc.edu/ns/ontology/intelligence\#$>$ . \\
@prefix rdf: $<$http://www.w3.org/1999/02/22-rdf-syntax-ns\#$>$ . \\
@prefix rdfs: $<$http://www.w3.org/2000/01/rdf-schema\#$>$ . \\
@prefix xml: $<$http://www.w3.org/XML/1998/namespace$>$ . \\
@prefix xsd: $<$http://www.w3.org/2001/XMLSchema\#$>$ . \\
@prefix dbp: $<$http://dbpedia.org/resource\#$>$ . \\
@prefix owl: $<$http://www.w3.org/2002/07/owl\#$>$ . \\
\\
$<$Int3482758232$>$ a intel:Intelligence ; \\
    intel:hasVulnerability $<$execute\_arbitrary\_code$>$ ;\\
    intel:hasVulnerability $<$denial\_of\_service$>$ .\\
\\
$<$crafted\_web\_site$>$ a uco:Means .\\
\\
$<$remote\_attackers$>$ a uco:Attacker .\\
\\
$<$Microsoft\_Internet\_Explorer$>$ a uco:Product ;\\
	uco:hasVulnerability $<$execute\_arbitrary\_code$>$ ;\\
    uco:hasVulnerability $<$denial\_of\_service$>$ ;\\
    owl:sameAs dbp:Internet\_Explorer .\\
\\
$<$execute\_arbitrary\_code$>$ a uco:Vulnerability ;\\
    uco:affectsProduct $<$Microsoft\_Internet\_Explorer$>$ ;\\
    uco:hasAttacker $<$remote\_attackers$>$ ;\\
    uco:hasMeans $<$crafted\_web\_site$>$ ;\\
    owl:sameAs dbp:Arbitrary\_code\_execution .\\
\\
$<$denial\_of\_service$>$ a uco:Vulnerability ;\\
    uco:affectsProduct $<$Microsoft\_Internet\_Explorer$>$ ;\\
    uco:hasAttacker $<$remote\_attackers$>$ ;\\
    uco:hasMeans $<$crafted\_web\_site$>$ ;\\
    owl:sameAs dbp:Denial-of-service\_attack .
\end{minipage}
\caption[RDF example.]{RDF for textual input ``Microsoft Internet Explorer 
allows remote attackers to execute arbitrary code or cause a denial of service (memory corruption) via a crafted web site, aka Internet Explorer Memory Corruption Vulnerability''. Also, $owl:sameAs$ property has been used to augment the data using an external source `DBpedia'.}
\label{fig:exampleRDF}
\end{figure}

\subsection{The VKG Structure}\label{model}

\begin{figure*}[ht]
\centering
\includegraphics[scale=0.3]{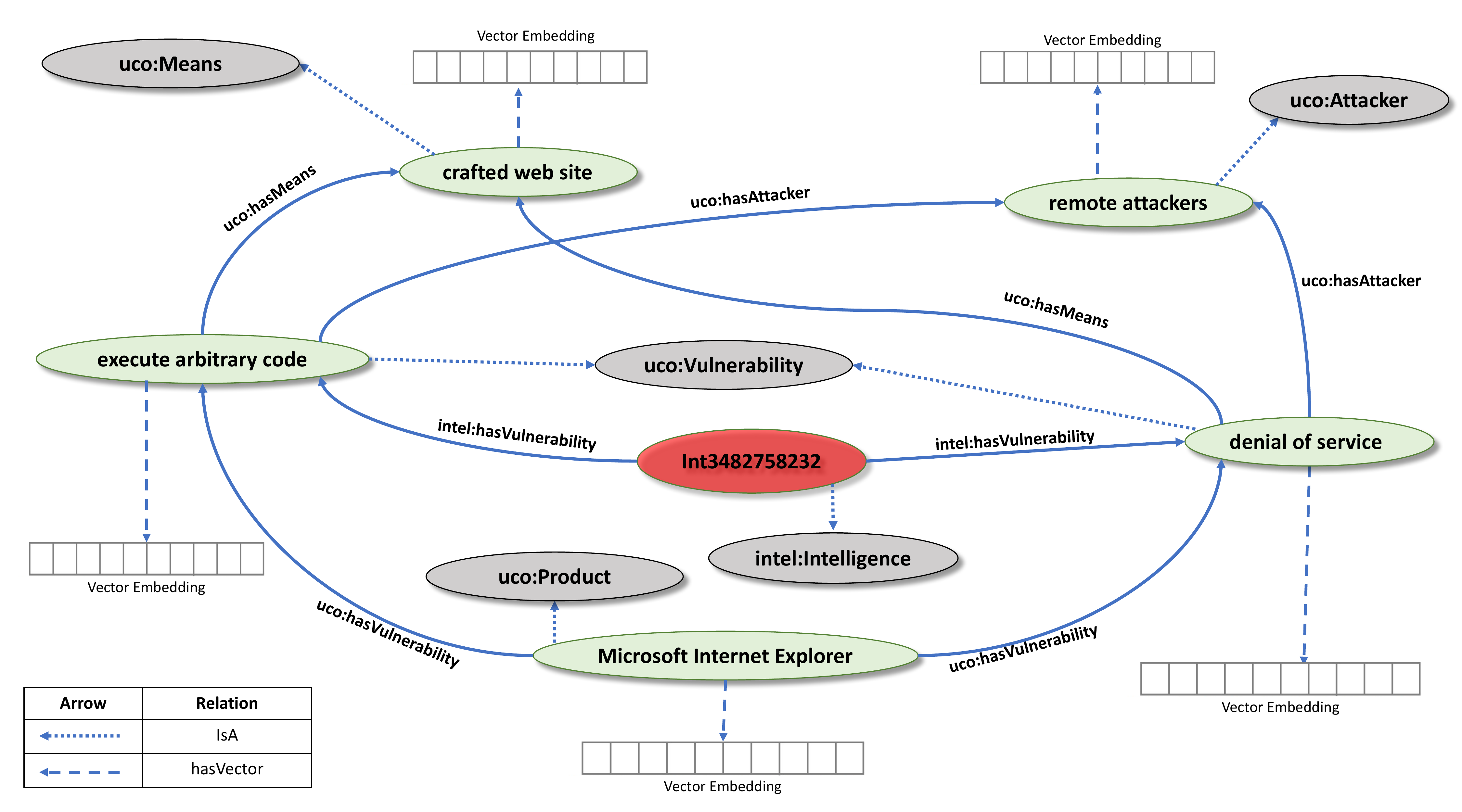}
\caption{In the VKG structure for \textit{``Microsoft Internet Explorer allows remote attackers to execute arbitrary code or cause a denial of service (memory corruption) via a crafted web site, aka ``Internet Explorer Memory Corruption Vulnerability.''} the knowledge graph part asserted using UCO includes the information that a product `Microsoft Internet Explorer' has vulnerabilities `execute arbitrary code' and `denial of service' that can be exploited by `remote attackers' using the means `crafted web site'. The knowledge graph entities are linked to their vector embeddings using the relation `hasVector'.}
\label{fig:cyberexample}
\end{figure*}


Here we describe our VKG structure, which leverages both vector spaces and knowledge graphs. 
In the VKG structure, an entity is represented as a node in a knowledge graph and is linked to its representation in a vector space. 
In an example (Figure \ref{fig:cyberexample}), entity nodes are linked to each other using explicit relations, as in a knowledge graph and are also linked to their word embeddings in a vector space. The assertions from Figure \ref{fig:exampleRDF} have been linked using the `hasVector' relation, to their embeddings. 

Now we discuss VKG population and various advantages it offers:

\subsubsection{Populating the VKG Structure} \label{pop}

In order to create the VKG structure, the structure population system requires as input, a text corpus. The aim of the system is to create the VKG structure for the concepts and entities present in the input corpus which requires us to create the knowledge graph and the vector parts separately and then linking the two. 


\begin{enumerate}

\item \textit{Creating semantic triples}:  An information extraction pipeline  extracts a knowledge graph from a collection of text documents. The first step applies components from Stanford CoreNLP components \cite{manning2014stanford} trained to recognize entities and relations in the cybersecurity domain to produce a knowledge graph for each document. 
The resulting knowledge graph is then materialized as an RDF graph.

\item \textit{Training vectors}: For the vector part of the VKG structure, we can generate entity embeddings using any of a number of vectorization algorithms (Section \ref{kgtovec}). For text, many of these are based on the `Relational Concurrence Assumption' principle \cite{mikolov2013distributed,mikolov2013efficient,pennington2014glove,huang2012improving}. 

\item \textit{Creating links between entity vectors and nodes}: We link knowledge graph nodes to their corresponding words in the vector space vocabulary using the $hasVector$ relationship (as shown in Figure \ref{fig:cyberexample}). Keeping the lexical tokens in the knowledge graph part allows us to update vector embeddings, if there is a need to refresh a stale vector model. Linking is initiated after the RDF triples are generated. 

\end{enumerate}




\subsubsection{Advantages and Agents}\label{adv}
The VKG structure helps us unify knowledge graphs and vector representation of entities, and allows us to develop powerful inference methods that combine their complementary strengths.

The vector representation we use, enable us to encode `local contextual knowledge'. These are based on the `Relational Concurrence Assumption' highlighted in \cite{mikolov2013distributed,mikolov2013efficient,collobert2011natural}. Word embeddings are able to capture different degrees of similarity between words. 
However, they are severely constrained while creating complex dependency relations and other logic based operations that are a forte of various semantic web based applications \cite{davis1993knowledge,berners2001semantic}. 

Knowledge graphs, on the other hand, are able to use powerful reasoning techniques to infer and materialize facts as explicit `global knowledge'. Those based on description logic representation frameworks like OWL, for example, can exploit axioms implicit in the graphs to compute logical relations like consistency, concept satisfiability, disjointness, and subsumption. As a result, they are generally much slower while handling operations like, ontology alignment, instance matching, and semantic search \cite{shvaiko2013ontology,JeanMary2009OntologyMW}. Knowledge graphs provide many reasoning tools including query languages like SPARQL \cite{sparql}, rule languages like SWRL \cite{swrl}, and description logic reasoners.


Potential applications that will work on our VKG structure, need to be designed to take advantages provided by integrating vector space models with a knowledge graph. In a general efficient use-case for our VKG structure, `fast' top-k search should be done on the vector space part aided by the knowledge graph, and the `slow' reasoning based computations should be performed on just the knowledge graph part. An input query can be decomposed into sub-queries which run on respective parts of the VKG structure (see \ref{query}).

Domain specific knowledge graphs are built using a schema that is generally curated by domain experts. When we link the nodes and embeddings we can use the explicit information present in these ontologies to provide domain understanding to embeddings in vector space. Adding domain knowledge to vector embeddings can further improve various applications built upon the structure. The vector embeddings can be used to train machine learning models for various tasks. 

Knowledge graph nodes in the VKG structure can be used to add information from other sources like, DBpedia, YAGO, and Freebase. This helps integrate information that is not present in the input corpus. For example, in Figure \ref{fig:cyberexample} we can link using the `$owl:sameAs$' property, `Microsoft\_Internet\_Explorer' to its DBpedia equivalent `dbp:Internet\_Explorer' \cite{owl}. Asserting this relation adds information like Internet Explorer is a product from Microsoft. This information may not have been present in the input cybersecurity corpus but is present in DBpedia.

Another advantage provided by integrating vector spaces and knowledge graphs is that we can use both of them to improve the results provided by either of the parts alone. For example (in Figure \ref{fig:cyberexample}), we can use the explicit information provided in the knowledge graph to aid the similarity search in vector space. If we are searching the vector space for entities similar to `denial\_of\_service', we can further improve our results by ensuring the entities returned belong to class `Vulnerability'. This information is available from the knowledge graph. This technique of knowledge graph aided vector space similarity search (\textit{VKG Search}, See Section \ref{query}) is used in our query engine. We execute similarity search on the embeddings and then filter out entities using the knowledge graph.

In Section \ref{appl}, we discuss applications we have created for the Cyber-All-Intel in detail.

\section{Cybersecurity Knowledge Improvement}\label{imp}
An added benefit of using multiple knowledge representation in the VKG structure is that we can use one representation to improve the other. Improving representation in turn improves the quality of applications that depend on them. In this section we will discuss how we can leverage the vector part of the VKG structure to improve the knowledge graph part and vice versa.
\subsection{Improving the Knowledge Graph using Vector Embeddings}\label{vectokg}
The vector representation of different entities can be used by a learning system to enhance the knowledge graph by predicting new relationships between entities. We have created a neural network that takes as input the vector representation of two entities and outputs the relation between them. 

This task will help in improving the knowledge graph as in when new data is added. An example task for this agent will be to predict the relation between the entities `android' and `buffer overflow', given data from a corpus that have both words. The triple can then be added to the knowledge graph along with their embeddings. 

\begin{figure}[ht]
\centering
\includegraphics[scale=0.27]{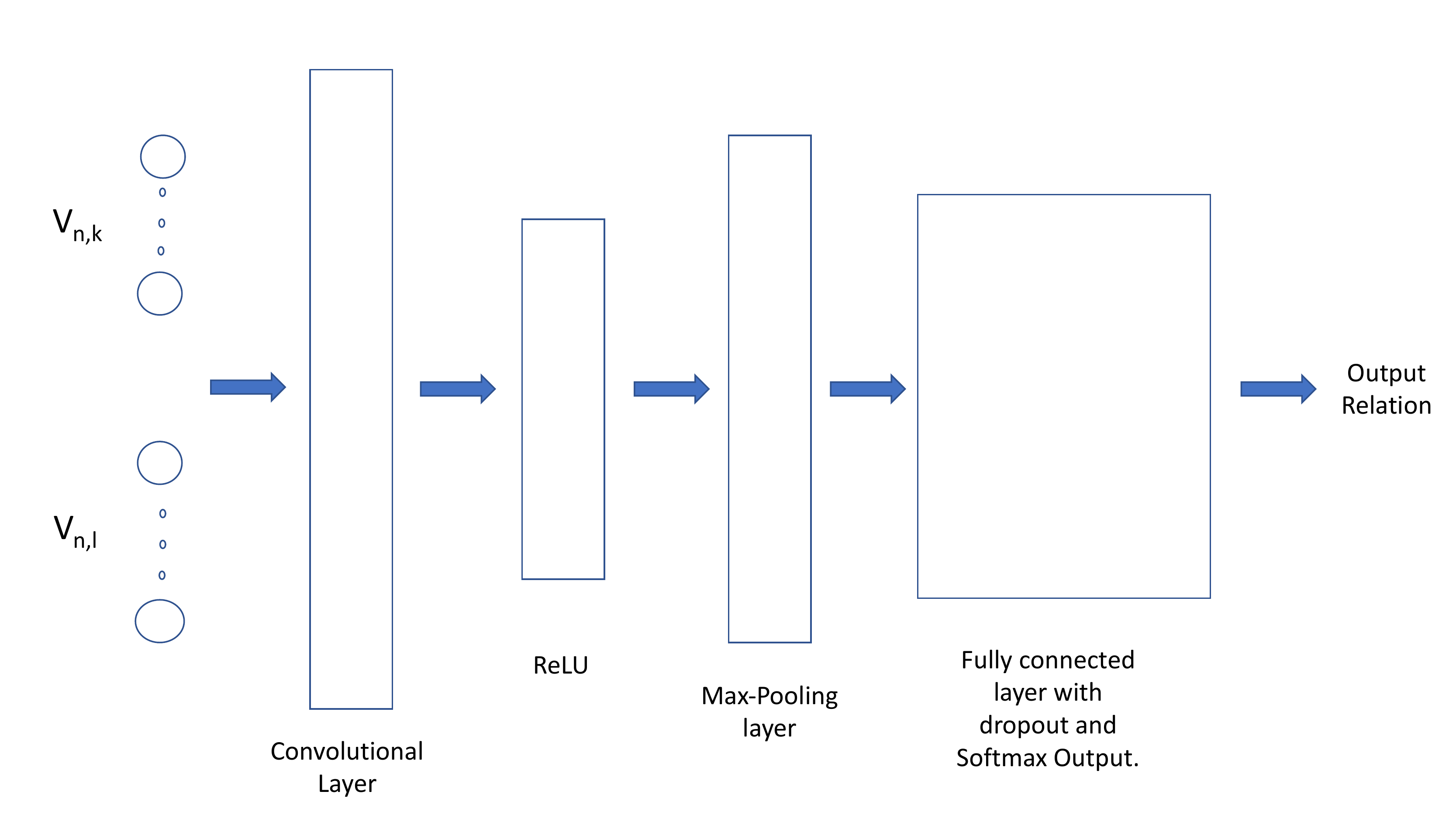}
\caption{Neural network structure to improve knowledge graph using embeddings.}
\label{fig:impk}
\end{figure}

Neural networks can be used to model nonlinear relations between inputs and outputs. We show the different layers of the neural network in Figure \ref{fig:impk}. The figure shows an input layer, multiple hidden layers and an output layer. The neural network has an input layer for embeddings followed by a convolutional layer. After this an activation function (ReLU layer) is added to introduce non-linearity, then a max-pooling layer for down-sampling, followed by a fully connected layer with dropout and softmax output.

Training this network is a supervised learning task. The training set ($TS$) includes:
\[ TS = \{(v_{1,1},\, v_{1,2}, R_{1}), (v_{2,1},\, v_{2,3}, R_{2}), ..., (v_{n,k},\, v_{n,l}, R_{m})\} \]
where, 
\[R_{1}, R_{2},..., R_{m} \in R\]
\[k,\, l \in E\]
\[ v_{n,k},\, v_{n,l} \in V  \, \& \, k \neq l \]

Here $R$ is the set of relations from the UCO ontology \cite{zareen_uco} from which the output is predicted. $R$ = [\textit{hasProduct, hasAttacker, hasMeans, hasConsequences, hasWeakness, isUnderAttack, hasVulnerability, ...}]. $V$ is the set of vector embeddings. $E$ is the set of entity classes. We ensure that input vectors are from different entity classes, i.e. $k, l$ are not same. 

Training this network involves minimizing the mean squared error function between the predicted value and the actual relation value. 

The triple generated using the two entities and the predicted relation is then added to the knowledge graph part of Cyber-All-Intel. We discuss the performance evaluations for this neural network in Section \ref{eval}.
\subsection{Improving Vector Embeddings using the Knowledge Graph}\label{kgtovec}
In this section we will discuss how we use the knowledge graph part of the VKG structure to improve the vector embeddings. The motivation for this task stems from the need to encode global context present as assertions in the knowledge graph, along with local co-occurring context in vector embeddings. Including the fact that `Samsung' and `Apple' are both mobile phone manufacturers in their vector embeddings will help bring these entities closer in the vector space. This will help improve various applications built upon the structure. The vector embeddings can be used to train machine learning models which will leverage the explicit assertions present in the knowledge graph part.

Current vector generation techniques proposed in \cite{mikolov2013efficient,pennington2014glove} provide a method to encode the relational concurrence in a vector space. However, these representations fail to include a global context. 
Ristoski et al. \cite{ristoski2016rdf2vec} created {\em RDF2Vec}, where they adapt neural language models for RDF graph embeddings. They transform the RDF graph data into sequences of entities, which are then considered as sentences. Using these sentences, they train the neural language models to represent each entity in the RDF graph as a vectors.

\begin{figure}[ht]
\centering
\includegraphics[scale=0.29]{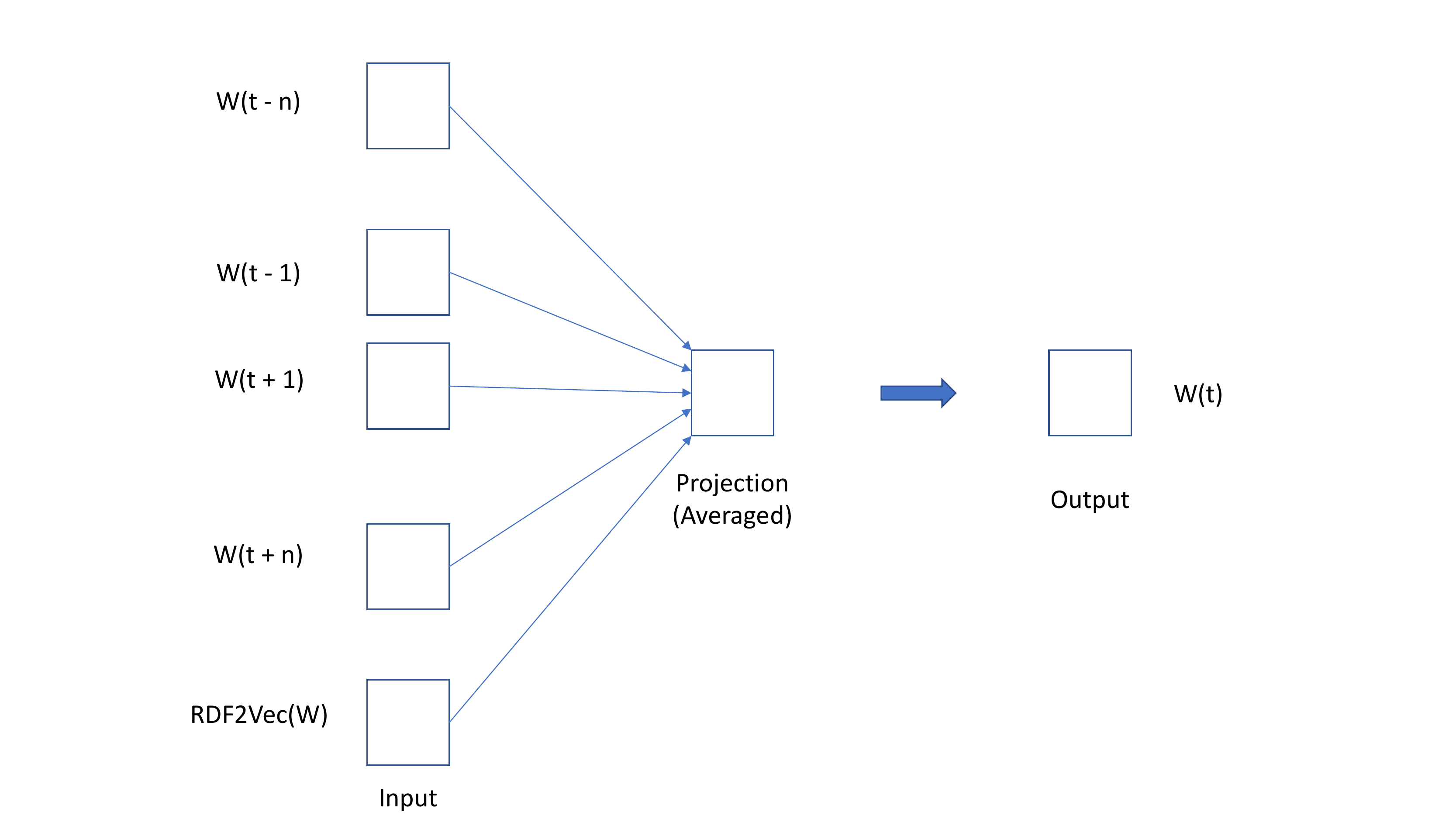}
\caption{Architecture for creating embeddings for $W(t)$ (Here, $t$ is the position of the word). Local context is provided by using co-occurring words $W(t-n)$,...$W(t-1)$, $W(t+1)$,...$W(t+n)$; $n$ is a hyper-parameter used for context window size. Global context is provided by the $RDF2Vec(W)$ vector created using the knowledge graph part.}
\label{fig:kgtov}
\end{figure}

In our approach, we created a feedforward neural network which takes as input the relational context of an entity, along with the RDF2Vec vector for that entity created using the knowledge graph part of the VKG structure. All the contextual words and the RDF2Vec vector get projected into the same position (as a result of vectors getting averaged). The training criterion is to correctly classify the current entity (i.e. the middle word). The output generated serves as the vector encoding for the entity. Figure \ref{fig:kgtov} shows the architecture of the neural network. We evaluate these vectors in Section \ref{eval}.

\section{Applications}\label{appl}
In this section we present two applications built using the VKG structure. The first one, allows a security analyst to issue complex query the VKG structure. The second one, generates alerts for the security analyst based on a `system profile'.
\subsection{A Query Processing Engine}\label{query}
An application running on the VKG structure described in, Section \ref{system} and populated via steps mentioned in Section \ref{pop}, can handle some specific type of queries. The application can ask a backend query processing engine to list declared
entities or relations, search for semantically similar concepts, and compute an output by reasoning over the stored data. This gives us three types of queries, $search$, $list$, and $infer$. These three are some of the basic tasks that an application running on the VKG structure will require, using which we derive our set of query commands ($C$):
\[ C = \{search,\, list,\, infer\} \]

A complex query posed by the application can be a union of some of these basic commands. An example query, to the Cyber-All-Intel security informatics system built on our VKG structure can be `list vulnerabilities in products similar to Google Chrome'; In this query we first have to $search$ for similar products to Google Chrome and then $list$ vulnerabilities found in these products. 


In the query mentioned above, $search$ queries, for the top-k nearest neighborhood search should be performed using the embeddings, and the $list$, $infer$ queries on the knowledge graph part. Domain experts can incorporate various reasoning and inference based techniques in the ontology for the knowledge graph part of the VKG structure. Mittal et al. in their system, CyberTwitter \cite{mittal2016cybertwitter} have showcased the use of an inference system to create threat alerts for cybersecurity using Twitter data. Such inference and reasoning tasks can be run on the knowledge graph part of the VKG structure.




For the Cyber-All-Intel system described in Section \ref{system}, some other example queries to the vector part can be, `Find products similar to Google Chrome.', `List vulnerabilities similar to buffer overflow', etc. We evaluate the performance of these queries on different parts in Section \ref{eval}.

\textit{Adding to SPARQL:} Our query processing engine aims at extending SPARQL. In SPARQL, users are able to write `key-value' queries to a database that is a set of `subject-predicate-object' triples. Possible set of queries to SPARQL are, $Select$, $Construct$, $Ask$, $Describe$, and various forms of $Update$ queries. We create a layer above SPARQL to help integrate vector embeddings using our VKG structure. Our query processing engine sends $search$ queries to the vector part of the structure, the $list$ query to the SPARQL engine for the knowledge graph, and the $infer$ query to the Apache Jena inference engine. Next, we go into the details of our backend query processing system.

\subsubsection{Query processing system} Let a query proposed by an application to the backend system on the VKG structure be represented by $Q^{VKG}$. The task of the query processing engine is to run the input query, $Q^{VKG}$, as efficiently as possible. We evaluate this claim of efficiency in Section \ref{eval}. We do not discuss a query execution plan as multiple expert plans can be generated by domain experts depending on the needs of the application.

As per our need, in the backend system, a query that runs only on the knowledge graph part and only the vector part of the structure are represented as $Q^{kg}$ and $Q^{v}$ respectively. An input query $Q^{VKG}$ can be decomposed to multiple components that can run on different parts namely the knowledge graph and the vector part: 
\[ Q^{VKG} \rightarrow Q^{kg} \cup Q^{v} \]

An input application query $Q^{VKG}$ can have multiple components that can run on the same part, for example, an input query can have three components, two of which run on the knowledge graph part and the remaining one runs on the vector part. Such a query can be represented as: 
\[ Q^{VKG} \rightarrow Q^{v} \cup Q_{1}^{kg} \cup Q_{2}^{kg} \tag{1}\label{q1}  \]

Where, $Q_{1}^{kg}$ and $Q_{2}^{kg}$ are the two components that run on the knowledge graph part and $Q^{v}$ component which runs on the vector part. 

It is the responsibility of the query processing system to execute these subqueries on different parts and combine their output to compute the answer to the original input query $Q^{VKG}$. We describe the query execution process using an example. 

\subsubsection{Example query} For the Cyber-All-Intel system an example query issued by the application: `Raise an alert if, a vulnerability similar to denial of service is listed in MySQL', can be considered as three sub-queries which need to be executed on different parts of the VKG structure.

The input query can be considered to be of the type \eqref{q1}, Where the subqueries are:
\begin{enumerate}
\item Finding similar vulnerabilities (set - $V$) to denial of service that will run on the vector embeddings ($Q^{v}$).
\item Listing known existing vulnerabilities (set - $K$) in MySQL ($Q_{1}^{kg}$).
\item Inferring if an alert should be raised if a vulnerability (from set $V$) is found in the product MySQL. This sub-query will run on the knowledge graph part ($Q_{2}^{kg}$).
\end{enumerate}
The query can be represented as: 
\begin{multline*}
Q^{VKG} = \{\{search, \, `denial\_of\_service', \, V\} \,\,\cup \\
\{list, vulnerability, `MySQL', \, K\} \,\,\cup \\
\{infer, V, K, `MySQL', alert\}\} \tag{Query 1}\label{q2} 
\end{multline*}

The query execution plan for \eqref{q2} is to first run $Q^{v}$ and $Q_{1}^{kg}$ simultaneously and compute the sets $V$ and $K$. After computing the sets the engine is supposed to run $Q_{2}^{kg}$.

The first part of the input query \eqref{q2}, is of the form $Q^{v}$ and will run on the vector part of the VKG structure. Its representation is: 
 \[ Q^{v} = \{search, \, `denial\_of\_service', \, V\} \]

\begin{figure}[ht]
\centering
\includegraphics[scale=0.2]{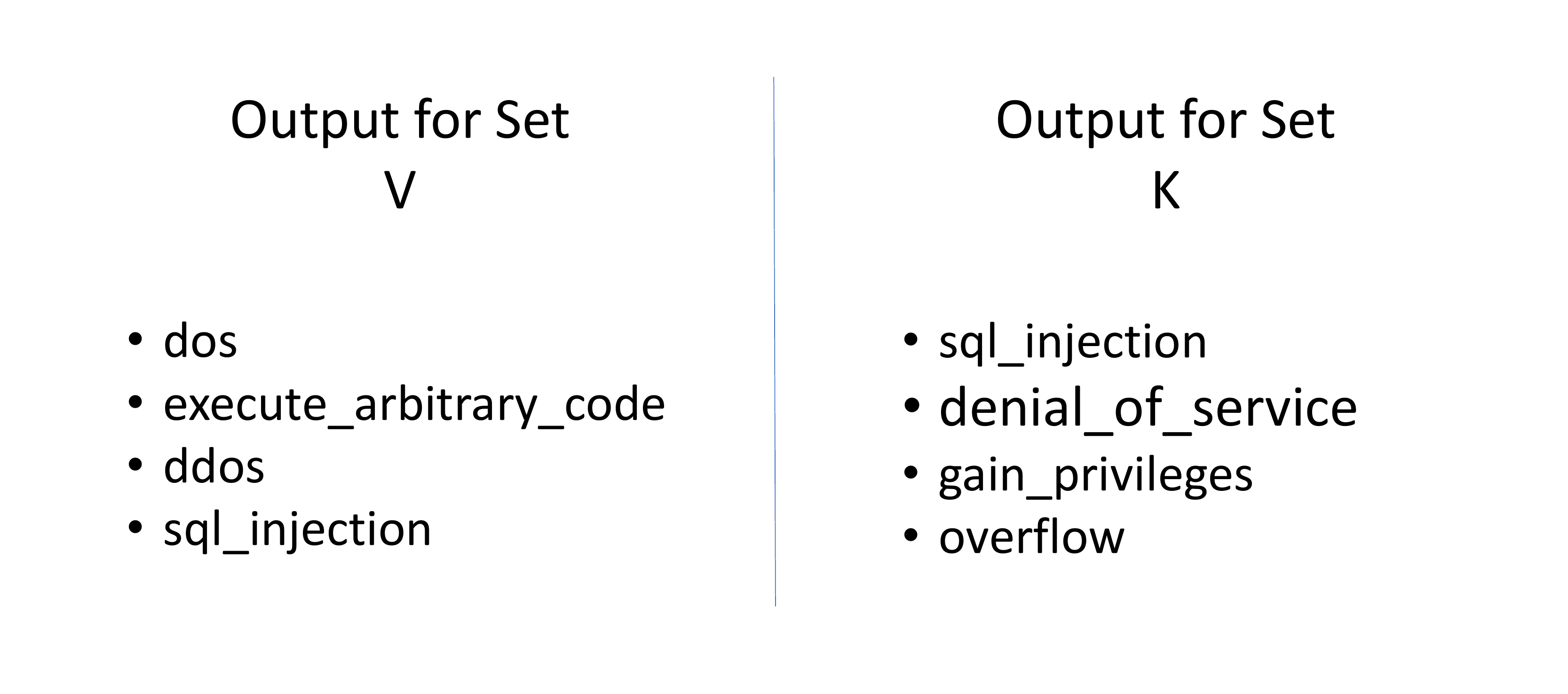}
\caption{The output of the sub-queries $Q^{v}$ and $Q_{1}^{kg}$ when run on the Cyber-All-Intel System. As there is some overlap between the sets $V$ and $K$ the output for the subquery $Q_{2}^{kg}$ will be `Alert = Yes'}
\label{fig:qo}
\end{figure}

The output generated is a set $V$ (Figure \ref{fig:qo}) and contains vulnerabilities similar to `denial\_of\_service'. We used the VKG search to compute this set and filter out all non vulnerabilities. The set $V$ will be utilized by other subqueries ($Q_{2}^{kg}$) to generate it's output. 

The second part of the input query \eqref{q2} is the first sub-query to run on the knowledge graph part of the VKG structure. 
\[ Q_{1}^{kg} = \{list, vulnerability, `MySQL', \, K\}\]

The goal of this query is to list all vulnerabilities (Figure \ref{fig:qo}) present in `MySQL' that are explicitly mentioned in the knowledge graph (set - $K$).

The third part of the input query \eqref{q2} is the second subquery to run on the knowledge graph part of the VKG structure. 
\[ Q_{2}^{kg} = \{infer, V, K, `MySQL', alert\} \]

Here, the output is to reason whether to raise an `alert' if some overlap is found between the sets $V$ \& $K$. Query $Q_{2}^{kg}$ requires an inference engine to output an alert based on some logic provided by domain experts or system security administrators. In Figure \ref{fig:qo} as there is overlap between the sets $V$ and $K$ an alert will be raised. 

\subsection{Knowledge Augmentation and Alerts}\label{rec}

In the field of cybersecurity a security analysts need to be aware of all possible threats and vulnerabilities to their cyber-infrastructure. We have created an intelligence alert system on top of the VKG structure, which briefs an analyst about various threats relevant to the software and hardware components present in an enterprise, when intelligence from multiple sources is analyzed and aggregated. 

In the past we created, \textit{CyberTwitter} \cite{mittal2016cybertwitter} to issue alerts about vulnerabilities found in various products used by an organization. The CyberTwitter system uses a knowledge graph where reasoning was done by adding SWRL rules. In the Cyber-All-Intel system, the SWRL rules have been extended and we also investigate similar products using the vector space to issue alerts based on an organization's `system profile'.

In this section we discuss two things, firstly, how we augment the knowledge graph with other sources of information and secondly, how we generate alerts.



\subsubsection{Knowledge Augmentation}\label{augment}

Many a times a query can come in that requires more knowledge for the answer to be computed than what is present in the text corpus used for training the VKG structure. An example query like ``What products similar to Internet Explorer are produced by Google Inc.?'' Such a query needs to first compute the set of possible products similar to `Internet Explorer' and then filter out the ones that are not produced by the entity `Google Inc.'. 

Knowledge graph nodes in the VKG structure can be used to add information from other sources like DBpedia \cite{auer2007dbpedia}, YAGO \cite{suchanek2007yago}, etc. This helps integrate information that is not present in the input corpus. Along with these sources we can add more information gathered from local organizational structure like network activity, shared library dependencies of a program executable, etc. This knowledge helps in adding local organizational knowledge to the system.

For our system to handle these type of queries we used the information present in existing knowledge graphs that were populated using other textual sources and techniques. As a proof of concept we integrated our \textit{Cyber-All-Intel} VKG structure's knowledge graph part with DBpedia \cite{auer2007dbpedia}. During this process of integration we asserted various products and vulnerabilities with their counterparts in the DBpedia knowledge graph. Figure \ref{fig:exampleRDF} shows an example where the property \textit{owl : sameAs} is used to assert counterparts for VKG instances of `Microsoft\_Internet\_Explorer', `Arbitrary\_code\_execution', and `Denial-of- service\_attack'. 

To include some local knowledge from the system that needs to be protected, we add shared library dependencies of programs installed. This information was collected using an Ubuntu system using the `objdump' tool\footnote{\url{https://sourceware.org/binutils/docs/binutils/objdump.html}} and filtering out library dependencies using the `NEEDED' flag. The dependencies for installed software were then asserted in a knowledge graph. 

Adding both global and local information to the knowledge graph helps us in improving the quality of alerts and recommendations.



\subsubsection{Generating Alerts}



For our alert system we create vector embeddings (for the VKG structure) using the augmented knowledge graph (with DBpedia and library dependencies of programs installed) discussed in Section \ref{augment}, and the generation method mentioned in Section \ref{kgtovec}. We extend the SWRL rules included in the CyberTwitter \cite{mittal2016cybertwitter} system. 


We also ask the security analyst to provide the recommender system a `system profile'. The profile contains information about the operating system, various installed softwares and their version information. We use the profile information as part of our rules to generate alerts.  

We created a VKG based alert system that has two logical parts: 
\begin{enumerate}
\item \textit{Vulnerability Alerts using factual data:} We created a rule based system to issue alerts using the knowledge graph part of the VKG structure which includes factual data like, collected intelligence, DBpedia, library dependencies of installed programs, etc. 

We utilized SWRL rules to include a reasoning engine analogous to a deductive based approach that a security analyst might take to figure out threats to her system. SWRL rules contain two parts, antecedent part (body), and a consequent (head). Informally, a rule may be read as meaning that if the antecedent holds (is ``true''), then the consequent must also hold. 

We have modified and generalized CyberTwitter's SWRL based recommendation engine \cite{mittal2016cybertwitter}, where we first compute if an intelligence is `valid and current' and in the second part we use a valid intelligence to raise an alert if its in the analyst's system profile. 

\item \textit{Vulnerability Alerts for similar products:} 
It is also necessary for the system to look for similar products that might also be at risk (The analyst can choose whether she wants these alerts). To keep a security analyst updated we also have to consider possible vulnerabilities that may exist in products that share library dependencies and/or developed by similar companies. For example, in case of products like Mozilla Firefox and Thunderbird, which are developed by the same company and have a considerable overlap in library dependencies; an alert generated for one, warrants an investigation into the other. 

So as to create such alerts we leverage the vector part of the VKG structure. When we get an alert about a vulnerability in a product from the factual data using the SWRL rules mentioned above, we look into possible intelligence obtained for products in the neighborhood of the vulnerable product and re-reason the SWRL rules with added information. We factor in the number of shared library dependencies, developing companies, etc in the SWRL rules. This vector neighborhood was created using the augmented knowledge graph mentioned in Section \ref{augment}. 


\end{enumerate}

After looking at the alerts generated using the factual data and by investigating similar products, we then push these alerts to the analyst depending on the organization’s `system profile’. 
We evaluate our recommender and alert system in Section \ref{eval}.


\section{Dataset and Experimental Setup}\label{setup}

For Cyber-All-Intel system, we created a Cybersecurity corpus as discussed in Section \ref{system} and shown in Figure \ref{fig:arch}. Data for the corpus is collected from many sources, including chat rooms, dark web, blogs, RSS feeds, social media, and vulnerability databases. The current corpus has 85,190 common vulnerabilities and exposures from the NVD dataset maintained by the MITRE corporation, 351,004 cleaned Tweets collected through the Twitter API, 25,146 Reddit and blog posts from sub-reddits like, r/cybersecurity/, r/netsec/, etc. and a few dark web posts \cite{dnmArchives}. 

For the vector space models, we created embeddings by setting vector dimensions as: 500, 1000, 1500, 1800, 2500 and term frequency as: 1, 2, 5, 8, 10 for each of the dimensions. The context window was set at 7. The knowledge graph part was created using the the steps mentioned in Section \ref{system} and the VKG structure was generated by linking the knowledge graph nodes with their equivalents in the vector model vocabulary (see Section \ref{pop}).

In order to conduct various evaluations, we first created an annotated test set. We selected some data from the cybersecurity corpus and had it annotated by a group of five graduate students familiar with cybersecurity concepts. The annotators were asked to go through the corpus and mark the following entity classes: \textit{Address, Attack / Incident, Attacker, Campaign, Attacker, CVE, Exploit, ExploitTarget, File, Hardware, Malware, Means, Consequence, NetworkState, Observable, Process, Product, Software, Source, System, Vulnerability, Weakness}, and \textit{VersionNumber}. They were also asked to annotate various relations including \textit{hasAffectedSoftware, hasAttacker, hasMeans, hasWeakness, isUnderAttack, hasSoftware, has CVE\_ID}, and \textit{hasVulnerability}. These classes and relations correspond to various classes and properties listed in the Unified Cybersecurity Ontology and the Intelligence Ontology \cite{mittal2016cybertwitter}. For the annotation experiment, we computed the inter-annotator agreement score using the Cohen's Kappa \cite{10.2307/2531300}. Only the annotations above the agreement score of 0.7 were kept.

The annotators were also tasked to create sets of similar products and vulnerabilities so as to test various aspects of the Cyber-All-Intel system. The most difficult task while designing various experiments and annotation tasks was to define the meaning of the word `similar'. Should similar products have the same vulnerabilities, or same use? In case of our cybersecurity corpus we found that the two sets, same vulnerabilities and same use were co-related. For example, if two products have SQL injection vulnerability we can say with certain confidence that they use some form of a database technology and may have similar features and use. If they have Cross-Site Request Forgery (CSRF) vulnerability they may generally belong to the product class of browsers.

Annotators manually created certain groups of products like, operating systems, browsers, databases, etc. OWASP\footnote{\url{https://www.owasp.org/index.php/Main_Page}} maintains groups of similar vulnerabilities\footnote{\url{https://www.owasp.org/index.php/Category:Vulnerability}} and attacks\footnote{\url{https://www.owasp.org/index.php/Category:Attack}}. We created 14 groups of similar vulnerabilities, 11 groups of similar attacks, 31 groups of similar products. A point to note here is that, in many cases certain entities are sometimes popularly referred by their abbreviations, we manually included abbreviations in these 56 groups. For example, we included DOS and CSRF which are popular abbreviations for Denial Of Service and Cross-Site Request Forgery respectively in various groups.

\section{Evaluations}\label{eval}

Cyber-All-Intel is a threat intelligence system which aims to provide tactical and operational support to the security analyst. The goal of the system is to add value to the analyst's work flow and enable her to make efficient security policy decisions. The system aims to reduce the `cognitive load' on the security analyst. To ensure that Cyber-All-Intel is able to sufficiently aid the analyst, we first evaluate the system by questioning it's core utilities. What is the quality of the information that is being provided by the system? Such an information can be about an attack or a vulnerability. 

The second method to evaluate our system is to measure how it can help a security analyst keep an updated policy for her organization. The system can provide the analyst with various similarities and differences between various variants of attacks. For example, the system can provide the analyst with the differences between `WannaCry' and `notPetya'. This information can then allow an analyst to create specific policy updates that help protect the organization. 

We also evaluate the knowledge improvement, query processing engine, alert generation capabilities of Cyber-All-Intel. 

\subsection{Evaluating core capabilities}

So as to evaluate the core capabilities of the system we focus on two features, first, the quality of new intelligence obtained or updates made to existing intelligences. Second, to evaluate if the system is able to highlight the similarities and differences between various attacks and vulnerabilities.

For the first one, we leverage the annotators mentioned in Section \ref{setup}. We provided them with the VKG structure generated along with the text that was used to generate it. For example, we provide an annotator with the VKG structure of `WannaCry' along with the text from our cybersecurity corpus that relates to WannaCry. The annotators were then asked to check if the VKG structure created was correct. 
We gave the annotators 60 such attacks, 49 of these were marked correct. Each attack was annotated by at least 2 annotators.

In the second one, we provided the annotators with pairs of attacks and vulnerabilities which are similar, as measured by comparing their VKG embeddings. We also gave them a policy to prevent one of the attacks. Their task was to modify the given policy so that it is able to protect an organization from both. For example, we provided our annotators with the VKG structure for `WannaCry' and `notPetya', along with a policy to prevent WannaCry. The annotators were then tasked to change the policy so that it can prevent both WannaCry and notPetya. We ran this experiment for 22 such pairs, along with the policies to prevent one of the attacks present in the pair. Each pair was annotated by at least 2 annotators. Of the given 22 pairs, the annotators were able to correctly modify 18 policies. 

Such capabilities add value to the analyst workflow. Providing this information can help the analyst make informed policy level changes. We would also like to bring to notice a possible feature, where an AI system will automatically suggest policy level changes to the security analyst. Research on this feature is ongoing but such a capability will be built using the core capabilities of the Cyber-All-Intel system. 

What is missing in existing proprietary SIEMs like LogRhythm, Splunk, IBM QRadar, and AlienVault, etc. is the integration of threat intelligence from disparate sources followed by efficient interpretation and reasoning on data using known intelligence \cite{netwrixlimits,netwrixinfo}. This can reduce false positives and improve the current state of the art in this domain. Also, it reduces the cognitive load on the analyst, because the system can fuse threat intelligence with observed data to detect attacks early, ideally left of exploit. 

\subsection{Evaluating knowledge improvement}
In Cyber-All-Intel we leverage different knowledge representations in our VKG structure. The knowledge graph part is designed to hold more global context. The vector space embeddings on the other hand have been created using the local context around the entity. In section \ref{imp}, we discuss our methodology to leverage the different parts of the VKG to improve each other. 

In the knowledge graph improvement task discussed in Section \ref{vectokg} we classify the relation between two entities using a deep learning model. The model was trained on the true positive relations explicitly declared in 60,000 CVEs\footnote{There were about 95,000 CVEs when this article was published.}. 30,000 were used as test set. Our model has an accuracy of 81.5\%. 

To evaluate the quality vector embeddings generated in Section \ref{kgtovec} we use the sets of `similar' products and vulnerabilities discussed in Section \ref{setup}. The task was to evaluate if similar real-word products and vulnerabilities are present in the same neighborhood in the vector space. For the task, we query the embeddings on one of the elements in the similar annotated sets and then compare the entities of the set returned (We compare products only to products, vulnerability only to vulnerability,...). For the vector space with dimensionality of 1500 and term frequency 2, we get a precision of 0.84 and recall of 0.24. 

\subsection{Evaluating the query processing engine}
Using the data and annotation test sets mentioned in Section \ref{setup}, we evaluate our query processing engine. In Section \ref{query}, we describe our query processing engine and its three query commands: $search$, $list$, and $infer$. Here we evaluate $search$ and $list$ query but not the $infer$ queries as they depend on the reasoning logic provided in the ontology and can vary with application.

\subsubsection{Evaluating the $search$ query}

\begin{figure}[ht]
\centering
\includegraphics[scale=0.29]{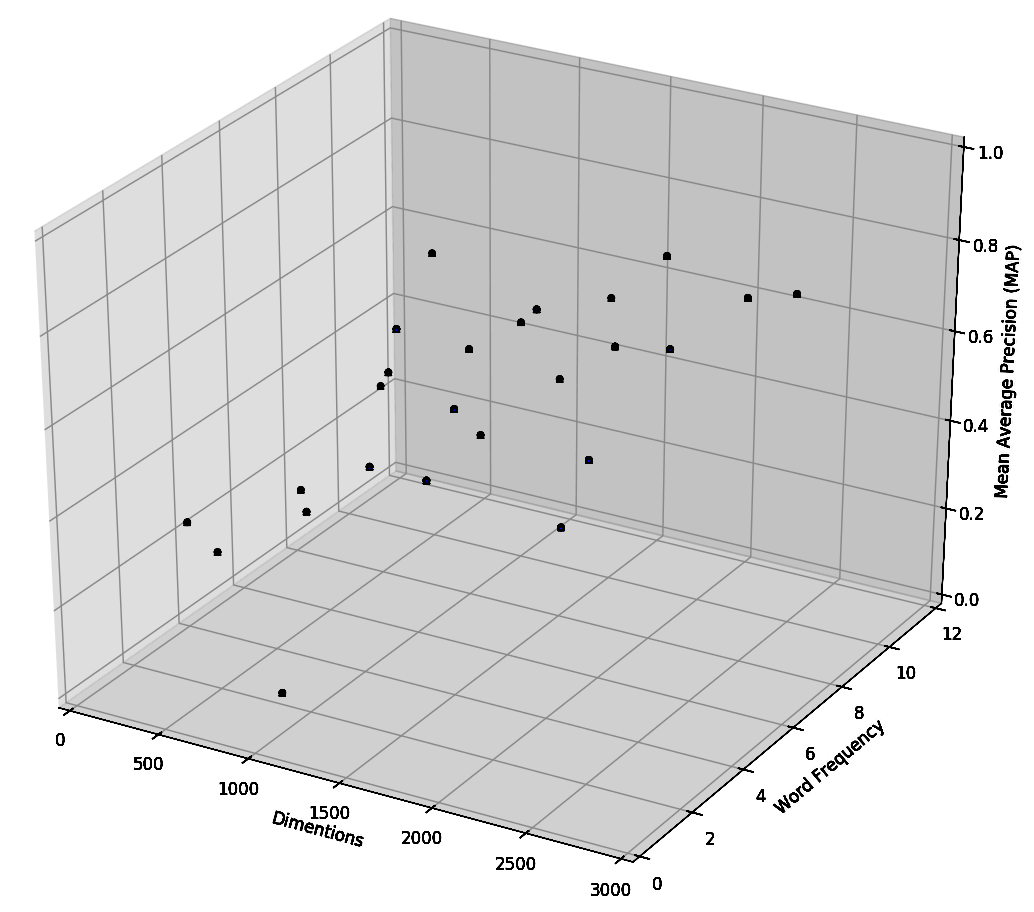}
\caption{Mean Average Precision for different dimensions and word frequency. Models with higher dimensions and word frequency performed better.}
\label{fig:map}
\end{figure}

An input query to the VKG structure to find similar concepts can either run on the vector space using various neighborhood search algorithms \cite{gionis1999similarity,Kuzi:2016:QEU:2983323.2983876} or the knowledge graph using ontology matching, ontology alignment, schema matching, instance matching, similarity search, etc. \cite{shvaiko2013ontology,DeVirgilio:2013:SMA:2457317.2457352,zheng2016semantic}. 
To evaluate the vector embeddings part of the VKG structure we used the `similar' sets created by the annotators. We trained various vector space models with vector dimension, 500, 1000, 1500, 1800, 2500 and term frequency, 1, 2, 5, 8, 10. Increasing the value of dimensionality and decreasing the term frequency almost exponentially increases the time to generate the vector space models. We first find the combination of parameters for which the Mean Average Precision (MAP) is highest, so as to use it in comparing the performance of vector space models with knowledge graphs and graph aided vectors in the VKG structure, in finding similar vulnerabilities, attacks, and products. For the 56 similar groups the vector model with dimensionality of 1500 and term frequency 2, had the highest MAP of 0.69 (Figure \ref{fig:map}). Models with higher dimensions and word frequency performed better.

To compare the performance of the $search$ query over vector space and its counterpart from the knowledge graph side we used the vector embedding model with dimensionality of 1500 and term frequency 2. To compute instance matching on knowledge graphs, we used an implementation of ASMOV (Automated Semantic Matching of Ontologies with Verification) \cite{JeanMary2009OntologyMW}.

\begin{figure}[ht]
\centering
\includegraphics[scale=0.25]{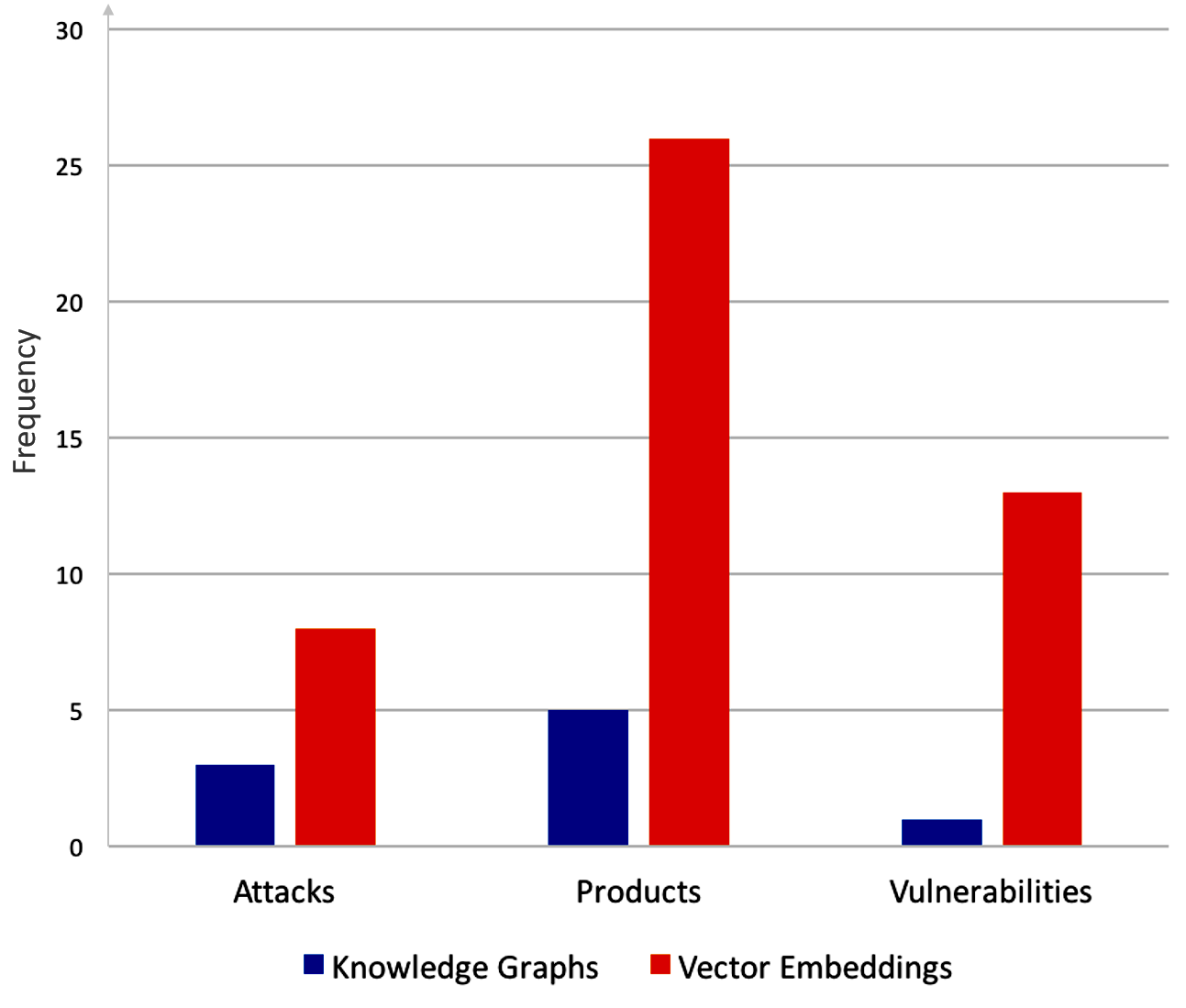}
\caption{The number of times the MAP score was higher for the two knowledge representation techniques for the 56 similar groups. Vector embeddings performed better than knowledge graphs. Embeddings performed better in 8 attacks, 26 products, and 13 vulnerabilities.}
\label{fig:map2}
\end{figure}

On computing the MAP for both vector embeddings and the knowledge graphs we found that embeddings constantly outperformed the knowledge graphs. Figure \ref{fig:map2}, shows that the MAP value for vector embeddings was higher 47 times out of 56 similarity groups considered. The knowledge graph performed significantly bad for vulnerabilities and attacks as the structural schema for both attacks and vulnerabilities was quite dense with high number of edges to different entities. This significantly affected the performance of schema matching.

To test the advantages of our VKG structure we evaluate the VKG search (see Section \ref{query}) against the vector space model. The VKG search on vector space achieved a MAP of 0.8, which was significantly better than the MAP score (0.69) achieved by using just the vector model. The reason for higher quality results obtained by using the VKG search is due to the fact that we can filter out entities by using class type declarations present in the knowledge graph.

\begin{table}[ht]
\centering
    \begin{tabular}{|c|c|c|c|}
    \hline
    \textbf{Model} & Graphs & Vectors & VKG Search \\ \hline
    \textbf{MAP}   & 0.43   & 0.69    & 0.80         \\ \hline
    \end{tabular}
    \caption{Best Mean Average Precision for knowledge graphs, vector space models, and VKG structure.}
\end{table}

\subsubsection{Evaluating the $list$ query}

Since the declarative assertions are made in the VKG's knowledge graph, to evaluate the $list$ query we evaluate the quality of the knowledge graph part. The $list$ query can not be executed on the vector space as there is no declarative information in embeddings. 

To check the quality of the knowledge graph triples generated from the raw text we asked the same set of annotators to manually evaluate the triples created and compare them with the original text. The annotators were given three options, correct, partially correct, and wrong. From 250 randomly selected text samples from the cybersecurity data, the annotators agreed that 83\% were marked correct, 9\% were partially correct, and 8\% were marked wrong.

\subsection{Evaluating the alert system}

In Section \ref{rec} we discussed our alert system. In our system we first generate alerts using the factual data obtained by augmenting incoming intelligence with shared library dependencies and DBpedia linkages. This data is then used by a rule based system to generate alerts. Once we get an alert about a product, we also investigate other products in it's vector neighborhood created using the augmented knowledge graph. Alerts are then pushed to the analyst depending on the organization's `system profile'. 

To evaluate the quality of these alerts we conducted a small user study where we asked five assessors to judge the usefulness of alerts (options: useful, maybe, useless) given the set of sources responsible for the alert. Out of 55 alerts generated 43 were marked as useful, 3 were marked useless, and the remaining 9 were marked as maybe.
\section{Conclusion \& Future Work}
\label{conc}

This paper presents \textit{Cyber-All-Intel} a system for knowledge extraction, representation and analytics in an end-to-end pipeline grounded in the cybersecurity informatics domain. The system creates a cybersecurity corpus by collecting threat and vulnerability intelligence from various textual sources like, national vulnerability databases, dark web vulnerability markets, social networks, blogs, etc. which are then represented as instances of our VKG structure. 

The Cyber-All-Intel system also pro-actively tries to improve the underlying cybersecurity knowledge. We have created neural network models, that take the vector part of the VKG structure and improves the knowledge graph. The knowledge graph part serves as the input to the vector generating part, adding more global knowledge to these embeddings.

We use the system to answer complex cybersecurity informatics queries and issue alerts to the system analyst. Some other applications that can be added in the future are: suggestions for policy updates, linking an organization's in-network and endpoint sensors to create a robust Intrusion Detection and Prevention System (IDPS), etc.
These extensions and planned future work, brings us closer to our main aim - creating an artificial intelligence system to aid the security analyst.

\section*{Acknowledgement}

The research was partially supported by a gift from IBM Research, Department of Defense (U.S.A), and MITRE.
\bibliographystyle{plain}

\bibliography{PHD,ref}
\end{document}